\documentclass[letterpaper, 10pt, conference]{ieeeconf}  % Comment this line out if you need a4paper

\IEEEoverridecommandlockouts                              % This command is only needed if 
\usepackage{booktabs}

\usepackage{cite}
\usepackage{amsmath,amssymb,amsfonts}
\usepackage{algorithmic}
\usepackage{graphicx}
\usepackage{textcomp}
\usepackage{xcolor}
\usepackage{subfig}

\usepackage{textcomp}

\usepackage{fancyhdr}
\usepackage[ruled,vlined]{algorithm2e}

\usepackage{amsmath}
\usepackage{fancyhdr,graphicx,amsmath,amssymb}
\usepackage[ruled,vlined]{algorithm2e}
\usepackage{bm}
\usepackage{xcolor}

\usepackage{url}
\usepackage{hyperref}
\hypersetup{
    colorlinks=true,
    linkcolor=blue,
    filecolor=magenta,      
    }

\usepackage{multirow} % Required for multirows
\let\labelindent\relax
\DeclareUnicodeCharacter{2212}{-}

\include{pythonlisting}

\def\BibTeX{{\rm B\kern-.05em{\sc i\kern-.025em b}\kern-.08em
    T\kern-.1667em\lower.7ex\hbox{E}\kern-.125emX}}

\usepackage{bm} % bold math
\usepackage{color} % change text color        

\usepackage{tikz}
\usetikzlibrary{decorations.pathmorphing} % for snake lines
\usetikzlibrary{matrix} % for block alignment
\usetikzlibrary{arrows} % for arrow heads
\usetikzlibrary{calc} % for manimulation of coordinates
\usepackage{tikzscale}

\tikzstyle{block} = [draw,rectangle,thick,minimum height=2em,minimum width=2em]
\tikzstyle{sum} = [draw,circle,inner sep=0mm,minimum size=2mm]
\tikzstyle{connector} = [->,thick]
\tikzstyle{line} = [thick]
\tikzstyle{branch} = [circle,inner sep=0pt,minimum size=1mm,fill=black,draw=black]
\tikzstyle{guide} = []
\tikzstyle{snakeline} = [connector, decorate, decoration={pre length=0.2cm,
                         post length=0.2cm, snake, amplitude=.4mm,
                         segment length=2mm},thick, magenta, ->]

\usepackage{etoolbox}
\makeatletter
\patchcmd{\@makecaption}
  {\scshape}
  {}
  {}
  {}
\makeatletter
\patchcmd{\@makecaption}
  {\\}
  {.\ }
  {}
  {}
\makeatother
\def\tablename{Table}
\def\tablename{Table}
\usepackage[compact]{titlesec}
\usepackage{enumitem}
\setlist{nolistsep,leftmargin=*}
\setlist{nolistsep}
\DeclareMathOperator*{\argmin}{arg\,min}

\usepackage{amsmath}
\usepackage{fancyhdr,graphicx,amsmath,amssymb}
\usepackage[ruled,vlined]{algorithm2e}

\usepackage{bm}
\usepackage{xcolor}
\usepackage[font=footnotesize]{caption}

\begin{document}
\include{pythonlisting}
\title{Unified Control Framework for Real-Time Interception and 
Obstacle Avoidance of Fast-Moving Objects with Diffusion Variational Autoencoder
}

% \title{Unified Control Framework using Diffusion Variational Autoencoder for Intercepting Flying Objects with Obstacle Avoidance}

% Unified Control Framework for Real-Time Interception and Obstacle Avoidance of Fast-Moving Objects by Robotic Arms in Dynamic Environments

\author{Apan Dastider, Hao Fang, and Mingjie Lin}

\maketitle
\begin{abstract}

Real-time interception of fast-moving objects by robotic arms in dynamic
environments poses a formidable challenge due to the need for rapid reaction
times, often within milliseconds, amidst dynamic obstacles. This paper
introduces a unified control framework to address the above challenge by
simultaneously intercepting dynamic objects and avoiding moving obstacles. Central to
our approach is using diffusion-based variational autoencoder for motion planning to perform both
object interception and obstacle avoidance. We begin by encoding the
high-dimensional temporal information from streaming events into a
two-dimensional latent manifold, enabling the discrimination between safe and
colliding trajectories, culminating in the construction of an offline densely
connected trajectory graph. Subsequently, we employ an extended Kalman filter to
achieve precise real-time tracking of the moving object. Leveraging a
graph-traversing strategy on the established offline dense graph, we generate
encoded robotic motor control commands. Finally, we decode these commands to
enable real-time motion of robotic motors, ensuring effective obstacle avoidance
and high interception accuracy of fast-moving objects. Experimental validation
on both computer simulations and autonomous 7-DoF robotic arms demonstrates the
efficacy of our proposed framework. Results indicate the capability of the
robotic manipulator to navigate around multiple obstacles of varying sizes and
shapes while successfully intercepting fast-moving objects thrown from different
angles by hand. Complete video demonstrations of our experiments can be found in
\url{https://sites.google.com/view/multirobotskill/home}.

\end{abstract}


\section{Introduction}

%The reaching problem is deeply rooted in the community of control and intelligent robot~\cite{Mohammed2021ComprehensiveRO}, including both theoretical part and application part The real-time reaching problem in the application of robotics is essential as many daily routines can be simplified as reaching tasks such as picking geometrically varying objects within time-limit and catching fragile dynamic objects with high precision, and such complex tasks require the intelligent robot to possess real-time reactions to the environments~\cite{dastider2023retro}. However, intercepting a flying object is different from many typical robotic reaching tasks that can be well studied in the laboratory or simulation settings~\cite{Mohammed2021ComprehensiveRO} and thus is believed as the hardest task in real-world robotic reaching problems. First, the flying object usually has a fast speed and nonlinear moving trajectory~\cite{objectsFlight, Marturi2018}. Second, the workspace is often cluttered with dynamic obstacles that are also moving and changing in the geometric shapes~\cite{Sanchez_review1_sensors}. Therefore, the above two difficulties result in the expensive cost to afford any \textit{perception latency}\cite{toofast}, i.e., performing real-time high-speed reaching of intelligent robots under the complicated environment with dynamic obstacles.

Intercepting flying objects presents unique challenges distinct from typical
robotic reaching tasks extensively studied in controlled laboratory
or simulation environments~\cite{Mohammed2021ComprehensiveRO}. 
Specifically, flying objects tend
to exhibit high speed and follow nonlinear trajectories~\cite{Mohammed2021ComprehensiveRO, dastider2023retro,objectsFlight,Marturi2018}. Additionally,
workspaces are often cluttered with obstacles that are constantly
moving and dynamically altering their geometric configurations~\cite{Sanchez_review1_sensors}. Consequently, existing methodologies often result 
in the prohibitively high cost of accommodating any
perceptual latency~\cite{toofast}. 
In other words, executing real-time, high-speed reaching
maneuvers with intelligent robots in complex environments laden with dynamic
obstacles becomes exceedingly challenging~\cite{toofast}.

\begin{figure}[htbp]                                       
    \centering                                             
    \includegraphics[width=1\linewidth]{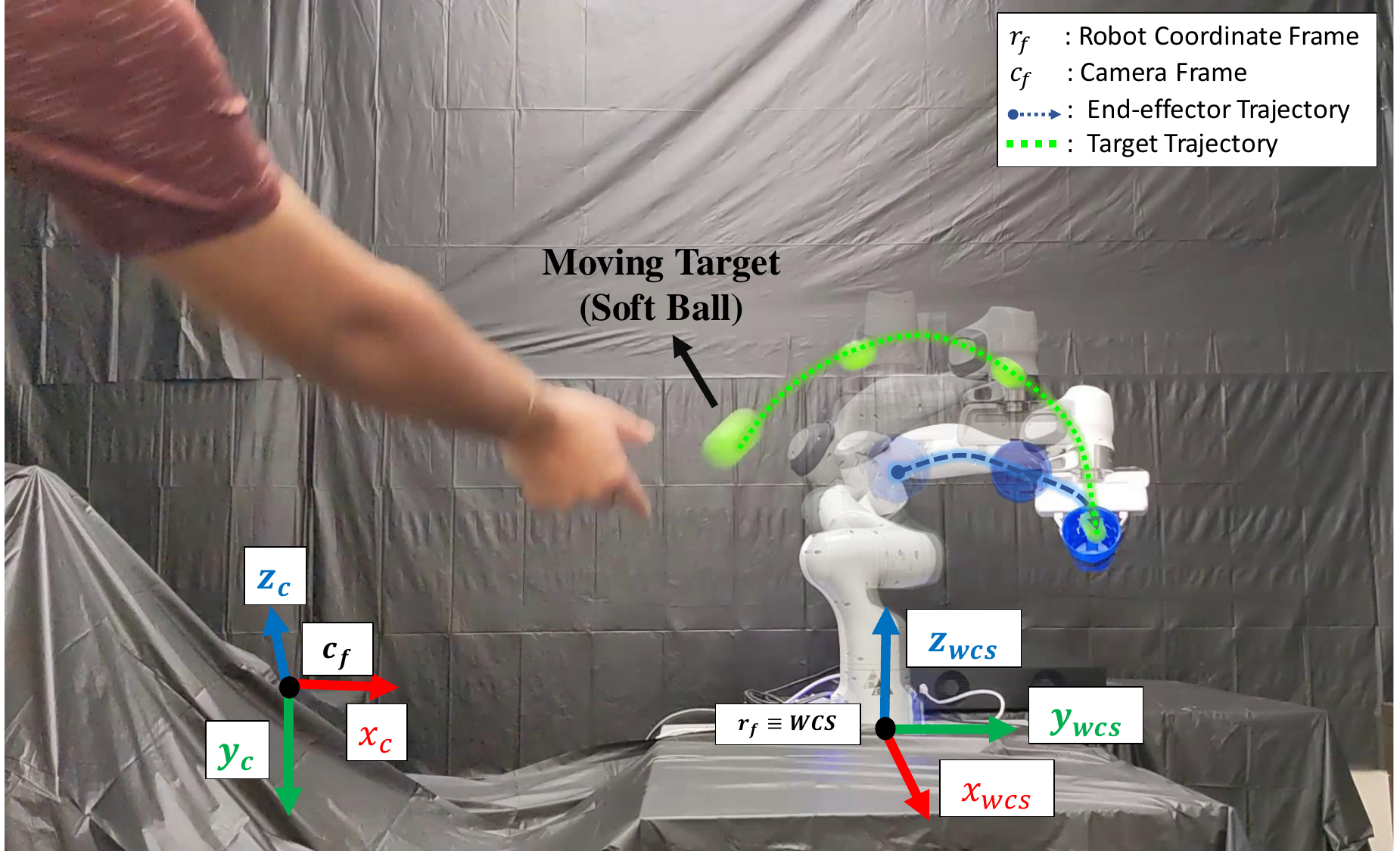}  
    \caption{Interception of a moving object by a 7-DoFs Robotic Manipulator.}
    \label{fig:platforms}                                   
\end{figure}
\begin{figure}[htbp]                                       
    \centering                                             
    \includegraphics[width=1\linewidth]{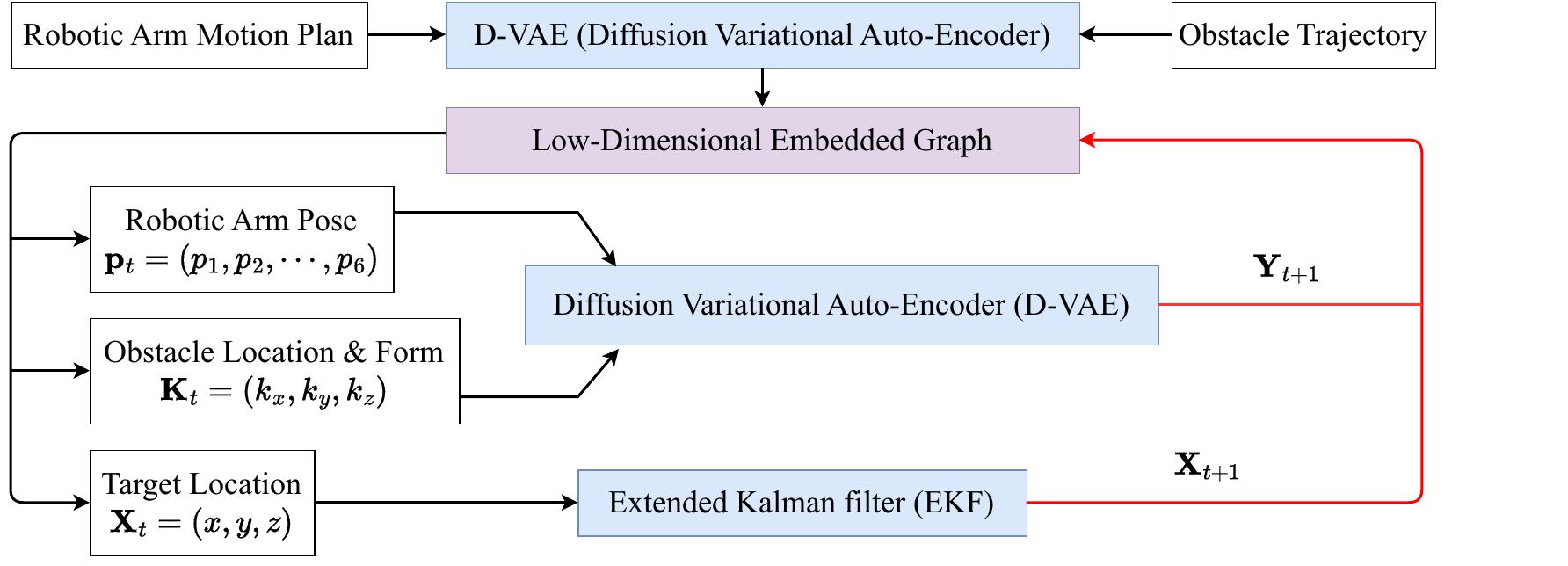}  
    \caption{Block diagram of our proposed unified control framework. More algorithm details can be found in Methods section Fig.~\ref{fig:method over all blocks}.}
    \label{fig:sys}                                   
\end{figure}

Numerous robotic motion planning algorithms have emerged in recent years, aiming
to enhance the efficacy of intercepting flying objects. Early algorithms such as
RRT \cite{LaValle1998RapidlyexploringRT, Salzman_rrt_IEEE_2016,
Zito_TwoLvlRRT_IROS_2012} were tailored for static objects in relatively simple,
obstacle-free environments. However, these approaches and their variants
\cite{ichter2019robot, mpnet, dynamicrrt, Wei_dynamic_sensors_2018} encounter
limitations in more complex obstacle environments. Traditional robotic
control algorithms resort to dynamical-system-based methods
\cite{khansari2012dynamical, stavridis_automation_2017}, wherein obstacles are
depicted as the workspace constraint, and later the optimal control theory is employed to
generate robot commands. Unfortunately, as workspace dimensions increase and
constraints become more nonlinear, solutions to optimal control problems
gradually become unattainable. Furthermore, extending traditional
dynamical-system-based methods to handle dynamical obstacles proves challenging,
necessitating the integration of time-varying workspace constraint into the
optimal control formulation\cite{raveendran2023dynamical}.

Recent advances in deep learning have facilitated the realization of flying
object interception and obstacle avoidance within an end-to-end learning
framework \cite{pmlr-v87-amiranashvili18a, hafner_corl_latent_dyn_2020,
Zhou2021}. A noteworthy innovation is the utilization of deep neural networks
for dimensionality reduction in the operational space, thereby expediting the
synthesis of reaching trajectories. For instance,
\cite{MohammadiHANR21} devised a path-planning algorithm on the latent workspace
manifold and introduced an obstacle avoidance scheme by modifying ambient
metrics. Motion planning networks (MPNet) \cite{mpnet} employ sampling-based
motion planning and obstacle avoidance via learned latent space networks. 
Recent work DAMON~\cite{Dastider_Damon_IROS_2023} leveraged the variational
autoencoder (VAE) structure to learn low-dimensional embeddings, upon which
motion trajectories were planned. However, relying solely on deep learning
techniques may not fully address interception challenges, particularly given the
high-speed motion of the target object. Firstly, the nonlinear and unpredictable
trajectory of the target object could confound purely deep learning-based
algorithms due to distribution shifts between training and test data
\cite{wu2022handling}. Secondly, re-synthesizing robotic commands based on
real-time object positions allows for only brief reaction times, potentially
insufficient for a single forward process in command synthesis \cite{rals}.
Consequently, there is a pressing need for more {\em unified} and {\em computationally efficient} robotic motion
planning algorithms to tackle these challenges effectively.

% In this paper, we proposed a unified control framework using diffusion Variational Autoencoder (DVAE) for dynamic object interception and collision avoidance. The major contributions of our can be highlighted as follows,
% \begin{itemize}
%    \item We develop a diffusion variational autoencoder (DVAE) to learn a two-dimensional latent manifold representation of the observed high-dimensional state vector, revealing the low-dimensional latent system dynamics of the robotic manipulator and environmental obstacles.
%    \item We construct a densely-connected graph network over the learned two-dimensional latent manifold to facilitate robotic motion planning with the shortest path routing by efficiently performing graph traversing on the \textit{collision-free} vertices.
%    \item We integrate the extended Kalman filter (EKF) for the accurate real-time estimation of the moving object in the reachable workspace of the robotic arm.
%    \item We decode the generation of encoded control commands for the real-time motion of robotic motors, which effectively guarantee the avoidance of moving obstacles and achieve high accuracy rates of intercepting a fast-moving object.
%\end{itemize}
% We demonstrated our proposed method in both computer simulation environments and real-world robotic arms. Our results hold promise for the future generalization of robotic motion planning algorithms to handle dynamic moving obstacles while intercepting a moving target in real time.

In this paper, we introduce a unified control framework utilizing diffusion variational autoencoder (D-VAE) for real-time dynamic object interception and collision avoidance (see Figure~\ref{fig:sys}). The significant contributions of our work are outlined as follows:

\begin{itemize}
\item
Development of a diffusion variational autoencoder (D-VAE) to model 
the complex high-dimensional state vector 
with
a two-dimensional latent manifold representation.  
This reveals the underlying low-dimensional dynamics of both the robotic manipulator and environmental obstacles. One unique aspect of our approach is that both robotic arm motion dynamics and moving obstacles are integrated into a unified model. 

\item
Construction of a densely-connected graph network over the learned latent manifold. This network facilitates efficient robotic motion planning by identifying the shortest path routes through graph traversing while ensuring collision-free motion.

\item
Integration of the extended Kalman filter (EKF) to provide accurate real-time estimation of moving objects within the reachable workspace of the robotic arm.

\item
Generation of decoded control commands to drive real-time motion of robotic motors. These commands effectively ensure avoidance of moving obstacles and achieve high accuracy in intercepting fast-moving objects.

\end{itemize}

We validate our proposed method through experiments conducted in both computer simulation environments and real-world scenarios with robotic arms. Our results demonstrate promising potential for the future generalization of robotic motion planning algorithms to effectively handle dynamic moving obstacles while intercepting moving targets in real time.

\section{Related Work}
\subsection{Robotic motion planning algorithm}
Current typical robotic motion planning algorithms include three perspectives: 1) RRT and its variants such as L2RRT and Dynamic RRT*~\cite{ichter2019robot,dynamicrrt}; 2) dynamical-system-based methods~\cite{khansari2012dynamical} 3) neural-network based approaches~\cite{dong2023review, Zhou2021}. For example, \cite{MohammadiHANR21} developed a path-planning algorithm on the latent workspace manifold and introduced an obstacle avoidance scheme by modifying the ambient metrics. Motion planning networks (MPNet)~\cite{mpnet} use sampling-based motion planning and obstacle avoidance through learned latent space networks.

\subsection{Variational autoencoder}
Variational autoencoder (VAE) is a powerful neural network architecture in many fields such as generative AI for robotics~\cite{ivanovic_gen_vae}, dimensionality reduction for computation efficient motion planning~\cite{Dastider_Damon_IROS_2023}, task conditioned movement leaning~\cite{pmlr-v100-noseworthy20a}. VAEs aim to optimize the network parameters by maximizing the evidence lower bound (ELBO) using the reparameterization trick. In our work, we augmented a VAE with techniques from the diffusion map operation for smooth robotic motion planning in low-dimensional manifold space.

\subsection{Kalman filter}
The Kalman filter, named by Rudolf E. Kalman, was developed back in 1961 and later was widely applied in modern society such as robotics~\cite{Urrea2021}, time series analysis~\cite{KF_time_series}, neural engineering~\cite{fang2023predictive,fang2023robust}. The basic idea of the Kalman filtering process is to use the linear control dynamical equations and sequences of output measurement to recursively estimate the underlying system states in real time. More recently, many variants of the Kalman filter were also proposed to deal with nonlinear control dynamics such as extended Kalman filter (EKF), unscented Kalman filter, and adaptive Kalman filter\cite{Khodarahmi2022}.

\section{Methods}
\begin{figure*}                                     
    \centering                                             
    \includegraphics[width=\linewidth]{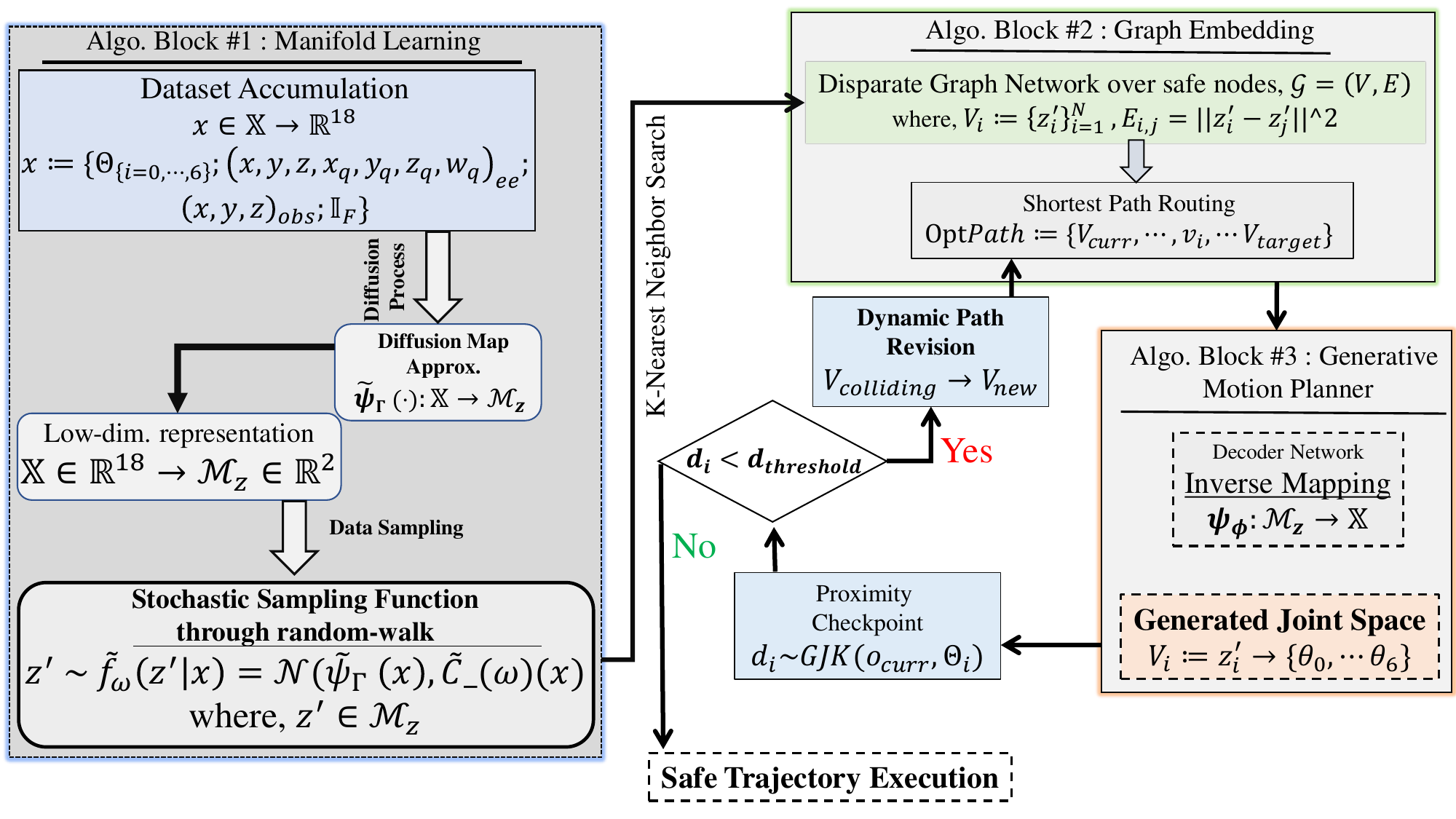} 
    \caption{Overall algorithm blocks of our proposed framework. We first encode the high-
dimensional data to a two-dimensional latent manifold using diffusion variational autoencoder (see left Block 1). Then. we construct an offline dense connected trajectory graph (see Block 2). We then leverage a graph-traversing strategy on the constructed offline dense graph to generate the encoded robotic motor control commands (also see Block 2). We last decode the generation of encoded control commands for the real-time motion of robotic motors (see Block 3).}% 
    \label{fig:method over all blocks}                                  
\end{figure*} 

Our method consists of three major components: a) a diffusion variational autoencoder to learn a two-dimensional latent manifold representation of the
observed high-dimensional robotic state vector (see section~\ref{Diffusion Variational Autoencoder}); b) a dense-connected graph network over the learned two-dimensional latent manifold to facilitate
robotic motion planning with the shortest path routing (see section~\ref{Dense-Connected Graph Network and Shortest Path Routing}); c) an extended Kalman filter (EKF) for the accurate real-time estimation of the moving object in the reachable workspace of the robotic arm (see section~\ref{Extended Kalman filter (EKF) for real-time estimation of the moving object}). Figure~\ref{fig:method over all blocks} depicts the comprehensive structure of our proposed method.

\subsection{Diffusion Variational Autoencoder}
\label{Diffusion Variational Autoencoder}
We develop a diffusion variational diffusion autoencoder (D-VAE) to efficiently fuse the advantage of variational autoencoder and diffusion maps for robot motion planning, which learns a two-dimensional latent manifold representation of the
observed high-dimensional robotic state vector. Consider a probability space $\mathbb{X}$ and random samples $\{x_i\}_{i=1}^N$ from the probability space ($N$ denotes the total number of samples), i.e., each $x_i \in \mathbb{X}$. Once we define Gaussian diffusion kernel $k_d(x_i,x_j)$, the D-VAE automatically encodes geometrical properties of the Euclidean space $\mathbb{X}$ into low-dimensional manifold space $\mathcal{M}_Z:=\widetilde{\psi}_\Gamma(\mathbb{X})$. 

\begin{figure}                                     
    \centering                                             
    \includegraphics[width=\linewidth]{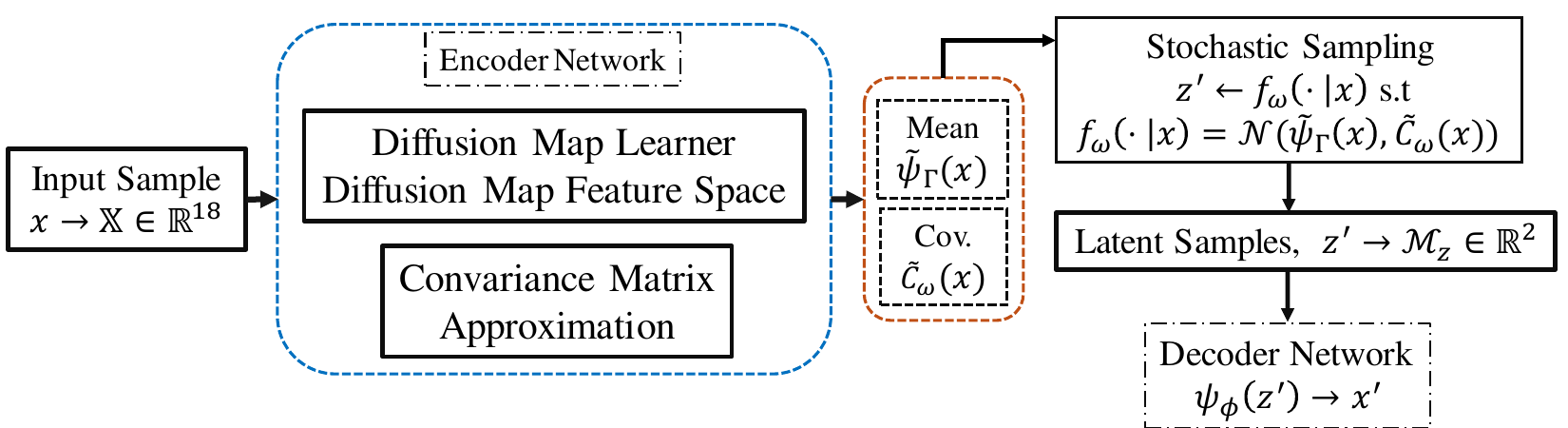}  
    \caption {The neural network architecture of our proposed diffusion variational encoder (D-VAE). The key difference of our proposed D-VAE with traditional VAE is the Encoder part, where we use the diffusion map to learn lower dimensional embedding efficiently. }
    \label{fig:method DVAE}                                   
\end{figure} 

% \textcolor{red}{please use equations and figures to formulate it, may be you can borrow the idea from Spectralnet.} 
The diffusion map learner is designed as a deep neural network-based approximator parameterized by $\Gamma$ to model the diffusion map feature space $\widetilde{\psi}_\Gamma(\cdot)$. Given the similarity affinity function between any $x_i$ and $x_j$, the loss function of the diffusion map learner is defined by,
\begin{equation}
    \mathcal{L}_\text{diffusion map} = \mathbb{E}[k_d(x_i,x_j)||z_i-z_j||^2_{2}]
\end{equation}
where, $z_i,z_j\in\mathcal{M}_Z$ and $\Gamma$ denotes the parameters for the mapping, $\mathcal{M}_Z:=\widetilde{\psi}_\Gamma(\mathbb{X})$. This loss function ensures that similar points in high-dimensional space are also in close proximity in lower-dimensional latent space.

Secondly, we learn a stochastic sampling function $\widetilde{f}_\omega(z^\prime|x)\rightarrow z^\prime$ which is defined by a Gaussian normal distribution, $\mathcal{N}(\widetilde{\psi}_\Gamma(x),\widetilde{C}_\omega(x))$ following \cite{Li2020VariationalDA}. Here, $\widetilde{C}_\omega(x)$ defines the covariance matrix of the random walk on manifold space $\mathcal{M}_Z$ for any sample $x_i$, and is parameterized by $\omega$. The inverse diffusion mapping is defined by $\psi_\phi:=\mathcal{M}_z\rightarrow\mathbb{X}$ and the learning parameter space for reconstruction is defined by $\phi$. Thus, the empirical loss function for D-VAE can be expressed as
\begin{equation}
    \mathcal{L}_\text{D-VAE}=-D_{KL}(f_\omega(z^\prime|x)||p_{\phi}(z^\prime|x))+\mathcal{L}_\text{decoder}.
\end{equation}
where the first term denotes the KL divergence from true diffusion probabilities and the second term defines the reconstruction error. Specifically, the decoder part for the inverse transformation $\psi_\phi:=\mathcal{M}_z\rightarrow\mathbb{X}$ is from the low-dimensional embedded space $\mathcal{M}_z$ back to the high-dimensional robotic operation space $\mathbb{X}$. To simplistically and effectively train the decoder model, we use the following L2-norm reconstruction error function
\begin{equation}
    \mathcal{L}_\text{decoder} \approx \underset{\omega, \phi}{\argmin} ||x - \psi_\phi(z^\prime)||^2_{2}
\end{equation}
where $z^\prime \sim \widetilde{f}_\omega(\cdot|x)$. Using this decoder, we essentially build a bijective mapping, where we can generate robotic motion control commands $\Theta^\prime_i$ through the low-dimensional embedding $z^\prime_i$. Using $\mathcal{L}_\text{diffusion map}$ and $\mathcal{L}_\text{D-VAE}$ together, we essentially learn a smooth latent manifold representation, where the original distance metric is preserved.
   
After we construct the architecture of D-VAE, we next apply it to reduce the dimension of our high-dimensional robotic data, which consists of 18 floating numbers. Specifically, each sample is 
defined by $[\theta_0, \cdots, \theta_6; \{x, y,z, x_q, y_q, z_q, w_q\}_{ee}; \{x,y,z\}_o; \mathbb{I}_F]$,
where $\theta_i$s determining the exact 
full joint-space pose of a robotic arm; $\{x, y,z, x_q, y_q, z_q, w_q\}_{ee}$ representing end-effector's coordinates and orientations in quaternion; $\{x, y, z\}_o$ defining the location of a point obstacle; $\mathbb{I}_F$ which is boolean collision flag. Essentially, the above 18 dimensions describe both the complicated robotic dynamics and geometrical properties of environmental obstacles. After sending high-dimensional robotic data into D-VAE, we stored the output low-dimensional embedded data $\{z^\prime_i\}_{i=1}^N \in \mathcal{M}_z$ in two-dimensional space by considering only the first two dominant components\cite{de2008introduction}, which later would be utilized to form a connected network in two-dimensional space for real-time motion planning (see next section).

\subsection{Densely-Connected Graph Network and Shortest Path Routing}
\label{Dense-Connected Graph Network and Shortest Path Routing}
Diffusion map-based graph construction and optimal
routing through connected graphs for real-world navigation and motion planning has been well established with a
view to learning obstacle avoidance \cite{hongDMRobot} and visual
navigation \cite{KupervasserDMRobot}. However, such graph-based planning has
been largely employed in mobile robots maneuvering on
planar Euclidean space.
Using the established D-VAE, we achieve the low-dimensional embedding via $\mathbb{X}\rightarrow\mathcal{M}_z$, where we can build graph $\mathcal{G}$ using ${z^\prime_i}$ in the embedding space $\mathcal{M}_z$ as
\begin{equation}
    \mathcal{G} = \text{graph}{(z^\prime_i)_{i=1}^N} \in \mathbb{R}^{2},
\end{equation}
which encodes the geometric properties of $\mathbb{R}^3\times\mathcal{S}^3$ space of a redundant robotic manipulator operating against point obstacles in $\mathbb{R}^3$. Therefore, graph $\mathcal{G} = (V, E)$ consists of nodes $V_{\{i=1,\cdots,N\}}\in\mathcal{M}_z$, which is associated with high-dimensional samples ${x_i} \in\mathbb{X}$; edge $E$ in the graph can be computed by Euclidean distance between two points in the low-dimensional space. Next, we develop a shortest path routing algorithm on the constructed graph $\mathcal{G} = (V, E)$. Notice that each low-dimensional node is labeled either as ``collision" or ``collision-free" based on the corresponding indicator $\mathbb{I}_F$. Thus, we apply the nearest neighbor search algorithm (greedy algorithm) on each ``collision-free'' node to perform the shortest routing algorithm for reaching a desired target node. We also adaptively update the previous routing to avoid any dynamic obstacles appearing in real time.

\subsection{Extended Kalman filter (EKF) for real-time estimation of the moving object}
\label{Extended Kalman filter (EKF) for real-time estimation of the moving object}
To predict the desired position of the target node, we use the well-known Extended Kalman filter (EKF) for real-time estimation of the moving object. Here, we consider a vector $s\in\mathbb{R}^n$ consisting of $n$ features on each frame captured using the image-based visual servoing system, where features can be determined as pixel coordinates $(f_x, f_y)$ on RGB frames and depth values $Z$ of respective coordinate from aligned depth frames. In detail, we have a bi-directional system integrating a eye-to-hand system and a hand-to-eye system. The interaction are described by the following dynamics, (also see supplementary materials in the attached link for more detailed derivations)
\begin{equation}
\label{eq: bi-directional dyanmics}
    \Dot{x} = 
    \begin{bmatrix}
    L_g(x_1,x_2){}^{c_f}\mathbb{T}_{r_f}J_{ee}u - v_o \\
    L_{Z_g}(x_1,x_2)({}^{c_f}\mathbb{T}_{r_f}J_{ee}u - L_g^\dag(x_1,x_2)v_o)
    \end{bmatrix} + w_t,
\end{equation}
where $x(t) = [s(t), z(t)] = [x_1(t),x_2(t)]$ where $s(t):=(f_x, f_y)$ and focal length, $\lambda$ of the vision sensor is known from the intrinsic properties of the sensor; control input, $u = \Dot{\Theta}$, hand-to-eye compatible depth Jacobian, $L_{Z_g}=-L_Z \mathbb{H}_{Rt}$ and $\Dot{s}_o = v_o$ the motion of feature calculated empirically from consecutive feature points and frame rate of the sensor. $L_g^\dag$ is the Moore-Penrose pseudo-inverse of the interaction matrix and $w_t \sim \mathcal{N}(0,Q_k)$ is additive Gaussian noise.  

Having the above explicit dynamics~\eqref{eq: bi-directional dyanmics}, in the \textit{prediction} phase of state estimation with known $f(x_{k|k},u_k)$ and its covariance matrix, $P_{k|k}$ at any time-step, $k$, we first required to linearize and discretize the system dynamics by first-order Taylor series approximation at each period $\Delta T$ by following Jacobians\cite{DVS, Julier1997NewEO} and updating the system dynamics and uncertainty matrix:
\begin{equation}
\begin{array}{lcl}
    \Bar{F} &=& \frac{\delta f(x,u)}{\delta x}|_{x=x_k, u = u_k} \\ 
    x_{k+1|k} &=& x_{k|k}+f(x_{k|k},u_k)\Delta T \\
    P_{k+1|k} &=& F_k P_{k|k} F_k^T + Q_k \;\text{where} \; F_k \approx \mathbb{I}+\Bar{F}\Delta T
\end{array}
\end{equation}
In our study, the measurements $y_k$ are image point features of dynamically moving objects in consecutive frames. When new measurements arrive, the $update$ phase improves the prediction outcomes of system dynamics through the following computations,
\begin{equation}
    \begin{array}{lcl}
            K_{k+1} &=& P_{k+1|k}C^T_{k+1}(C_{k+1}P_{k+1|k}C^T_{k+1}+R_{k+1}) \\
    x_{k+1|k+1} &=& x_{k+1|k}+K_{k+1}(y_k-C_{k+1}x_{k+1}|k) \\
    P_{k+1|k+1} &=& (\mathbb{I}-K_{k+1}C_{k+1})P_{k+1|k} 
    \end{array}
\end{equation}
where, $C_k$ is the measurement sensitivity matrix and $K_k$ is the Kalman gain. Once the filter converges, we can accurately predict the target's future state. Also, the coordinate values in $(x,y)$ plane of $c_f$ can be computed by using the projection model $(f_x,f_y) \rightarrow (X = f_x Z / \lambda,Y = f_y Z / \lambda)$. Therefore, we can in real-time estimate the future 3D coordinates of target location in robot's coordinate frame by ${}^{r_f}\mathbb{T}_{c_f}(x,y,z)_{c_f}$.

To this end, we complete the design of our algorithm. A summary of the proposed algorithm can be found in the algorithm~\ref{Algorithm 1} and figure~\ref{fig:method over all blocks}.
\begin{algorithm}    
	\caption{{Unified Control Algorithm}}
	\label{Algorithm 1}
		\textbf{Input:} \\
            Threshold Obstacle Distance $= \lambda_{obs}$ \\
            Threshold Target Pose Change $=\lambda_{obj}$
            \\Initial State $=x_o$ \\
            Current Goal State $=x_{g}$ \\
            Connected Graph Network, $G=(V,E)$
            
            \textbf{Output:}
            \\Optimal Control Sequence, $\Theta^*$\\
  \SetKwProg{Fn}{Function}{}{}\Fn {\text{Unified Control Algorithm($x_o, x_g, \lambda_{obs}, \lambda_{obj}, G$):}}{

        \texttt{GET} : 
        
        Initial Node on $G$, $V_o\leftarrow \widetilde{f}_\omega(\cdot|x_o)$ \\
        Current Goal Node on $G$, $V_g\leftarrow \widetilde{f}_\omega(\cdot|x_g)$ \\
        Initial Path, $P_{opt}\leftarrow GraphTraverse(V_o,V_g)$ \\
        \While{not intercepted or $t\leq t_{allowed}$}{
        \texttt{CALCULATE:}\\
        Distance, $d_{obs}\gets GJK(robot,obstacle)$ \\
        Estimated Target Pose, $O_{curr}\gets EKF(\cdot)$

        \eIf{$d_{obs}<\lambda_{obs} \; or \; ||O_{curr}-O_{old}||_2>\lambda_{obj}$}{
        \texttt{UPDATE:}\\
        Replanned Path, $p_{opt}^\prime\gets GraphTraverse(V_{curr},V_g^{new})$\\
        }
        { \texttt{EXECUTE:}\\
        Inverse Mapping, $\mathcal{M}_z\rightarrow\mathbb{X} \sim V_i \rightarrow x_i$ \\
        Robot Control , $\Theta_i\rightarrow \{ \theta_0,\cdots,\theta_6\}$
        }
        }
		% \For{$t \gets 1$ to $T$} {
  %       \texttt{GET} : $\{obs\}$ of target function \\
  %       \texttt{UPDATE} : Calc. Posterior $P^\prime(o_t|obs)$ \\
  %       \texttt{CALCULATE} : $D_{KL}(P^\prime(o_t|obs)||P(o_t))$
  %       \\\uIf{$D_{KL}(\cdot||\cdot)>\lambda_{thres}$}{
  %           \texttt{CALCULATE} : $z(o_t)$, $\delta u_t^*$ \\
  %           \texttt{FINETUNE} : $u_{t}^{*}\gets\Bar{u}_{t}+\delta u_{t}^{*}$
        
  %       }
  %       \texttt{EXECUTE} : $u_{t}^{*}$
    % }
} 
\end{algorithm}
\begin{figure*}                                     
    \centering                                             
    \includegraphics[width=\linewidth]{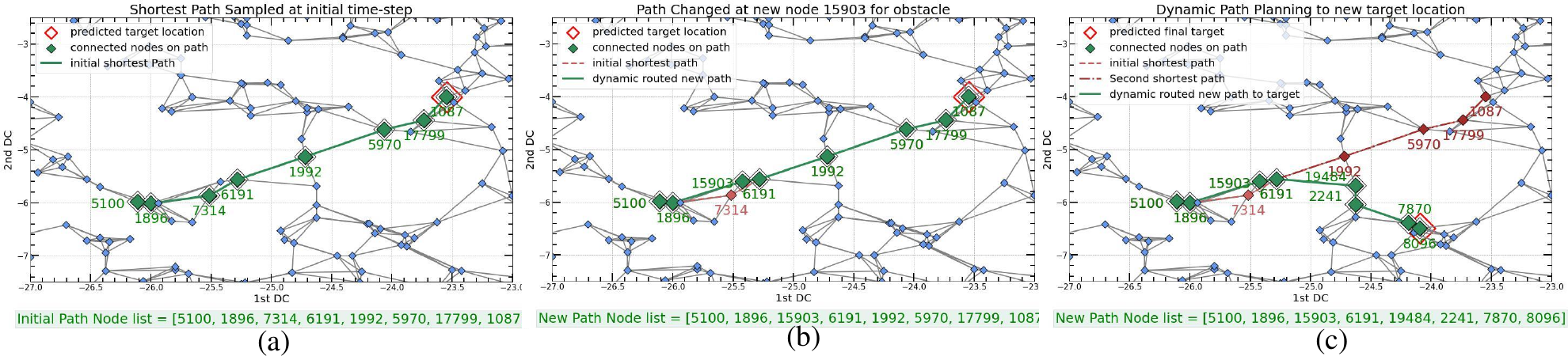}  
    \caption{Dynamic Routing on graph. (a) Initial trajectory (b) New trajectory to avoid obstacles, (c) Dynamic trajectory revising to intercept objects.}% 
    \label{fig:dynamic_adaption}
\end{figure*} 
\subsection{Evaluation metrics}
To effectively evaluate the performance of our proposed method, we define the following evaluation metrics. First, we use the successful ratio to measure the completeness of intercepting flying objects as 
\begin{equation}
    R_s = \frac{S}{N} \times 100\%,
\end{equation}
where $S$ counts the successful tasks and $N$ denotes the total number of tasks that we performed. Next, we record the experimental time $T$ to indicate the computational costs for successfully achieving tasks. Last, we use the $L_2$ distance to quantify the difference between the oracle trajectory of the target moving objects, $\tau_{oracle}$ and our prediction trajectory of the target moving objects. $\tau_{pred}$ to prove the efficiency of EKF,
\begin{equation}
    D = ||\tau_\text{oracle} - \tau_\text{pred}||_{2}.
\end{equation}

\section{Platform Overview}
\label{sec:system}
\subsubsection{Simulation platform}
To learn the low-dimensional embedding, we prepare the training dataset using the simulation model of 7-DoFs Franka Emika Panda robotic arm inside the simulation platform of Python Robotics Toolbox (RTB)\cite{rtb}. 
% \textcolor{red}{Please add a description for the RTB systems. 2-3 sentences are enough}
RTB was recently released as a variant of the famous robotics toolbox for Matlab and provides advanced API functionalities such as collision checking and distance calculation among meshes for efficient data-collection procedures and easy validation tasks.
Each collected sample has $18$ numerical floating values, lying in $18$ dimensional space--$7$ joint angles $\{\theta_i\}_{i=0,..,6}$, $7$ pose variables of end-effector containing pose coordinates and quaternion orientations in $\mathbb{R}^3\times\mathcal{S}^3$ space, location of point obstacle $\{x,y,z\}_{o}$ and collision flag $\mathbb{I}_F$. Next, we sent uniformly generated random control signals within the safety ranges of joint angles to the simulation model to exclude data bias and ensure enough data variance in preparing training dataset. The obstacle locations are also generated randomly via binary distribution to maximize numbers of ``\textit{collision and collision-free}'' samples. To have a dense sampling of the manipulator's operation space, we gather in total of 200k samples in our dataset for training the diffusion variational autoencoder. We assumed that by dense sampling, a collision-free trajectory for reaching the target object location is feasible. 

\subsubsection{Hardware platform}
For the demonstrations on the real-hardware platform, we utilize the 7-DoF Franka Emika Panda robotic manipulator (see Figure~\ref{fig:platforms}). The base link of the robot arm is fixed on the top of the table. Our object tracking and pose estimation algorithm, and adaptive trajectory planning pipeline run on a QUAD GPU server equipped with the Intel Core-i9-9820X processor through straightforward Python implementation. For obstacle detection and localizing dynamic target in $\mathbb{R}^3$ space, we incorporate Intel RealSense Depth Camera D435i and extracted the depth information to obtain coordinates $\{x, y, z\}_{target}$ in a three-dimensional camera coordinate frame. We also track the current obstacle location through depth sensing and extract the geometry of the obstacle with a safety barrier around it. Instantly, we transform this location into the robot's coordinate system to catch the thrown target smoothly and timely. These continuous transformations and feedbacks from depth sensing expedited the optimum routing on the $\mathbb{R}^2$ manifold graph for parallel smooth execution of obstacle avoidance and target reach task. To facilitate low-latency data communication between the actual controller and Franka controller interface, we design the whole system inside the Robot Operating System (ROS) using the Franka Integrated Library -- Libfranka and Franka ROS.

\section{Results}

We use comprehensive experiments in both computer simulations and hardware platforms to evaluate our proposed method. Specifically, our experiments seek to investigate the following questions: 
\begin{enumerate}
\item Can our unified control framework enable the robotic manipulator to smoothly avoid obstacles and reach a target object location in simulation platforms?  
\item How is the flying object intercepting performance in terms of successful rate compared to the state-of-the-art robotic motion planning algorithms?  
\item Can our diffusion variational autoencoder learn a lower-dimensional manifold such as $\mathbb{R}^2$ while remaining disparity between the collision-free and collision samples?      \item Can we concurrently localize the fast-flying object through image-based visual servoing on consecutive frames and robustly predict the future position using the proposed extended Kalman filter (EKF)? 
\end{enumerate}

%%%%%%%%%%%%%%%%%%%%%%%%%%%%%%%%%%%%%%%%%%%%%%%
\subsection{Intercepting flying objects with obstacle avoidance in both simulation and hardware platforms}
\begin{figure}                                       
    \centering                                             
    \includegraphics[width=\linewidth]{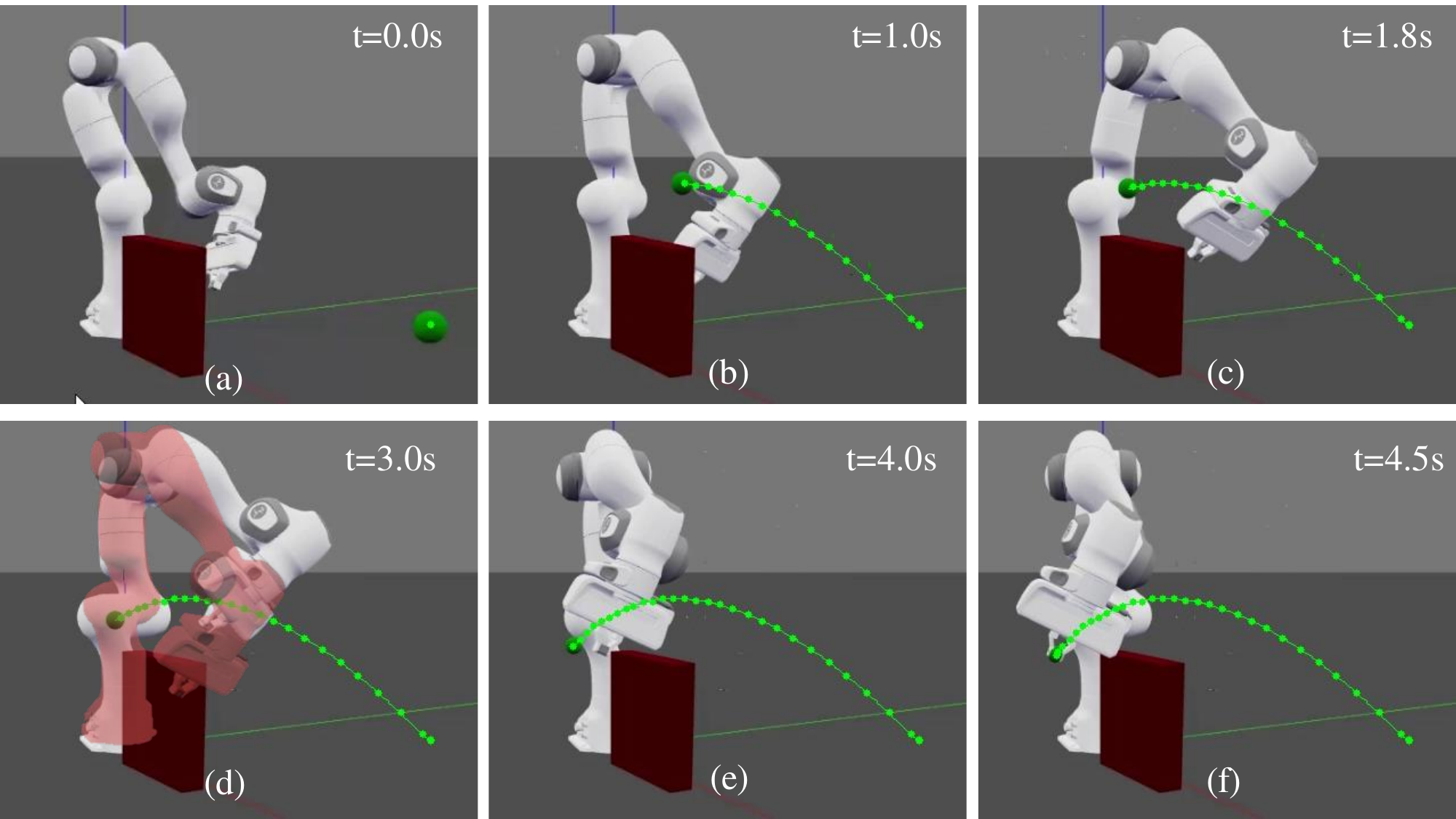}  
    \caption{ (a) Initial configuration. (b)-(c) Robot initializes to move to the predicted location, (d) Robot reroutes through a new path over the red obstacles, (in red shaded -- collision with obstacles without real-time planning), (e)-(f) Robot catches the thrown ball at the final target location.}% 
    \label{fig:software sim}                                   
\end{figure}

\begin{figure}                                     
    \centering                                             
    \includegraphics[width=\linewidth]{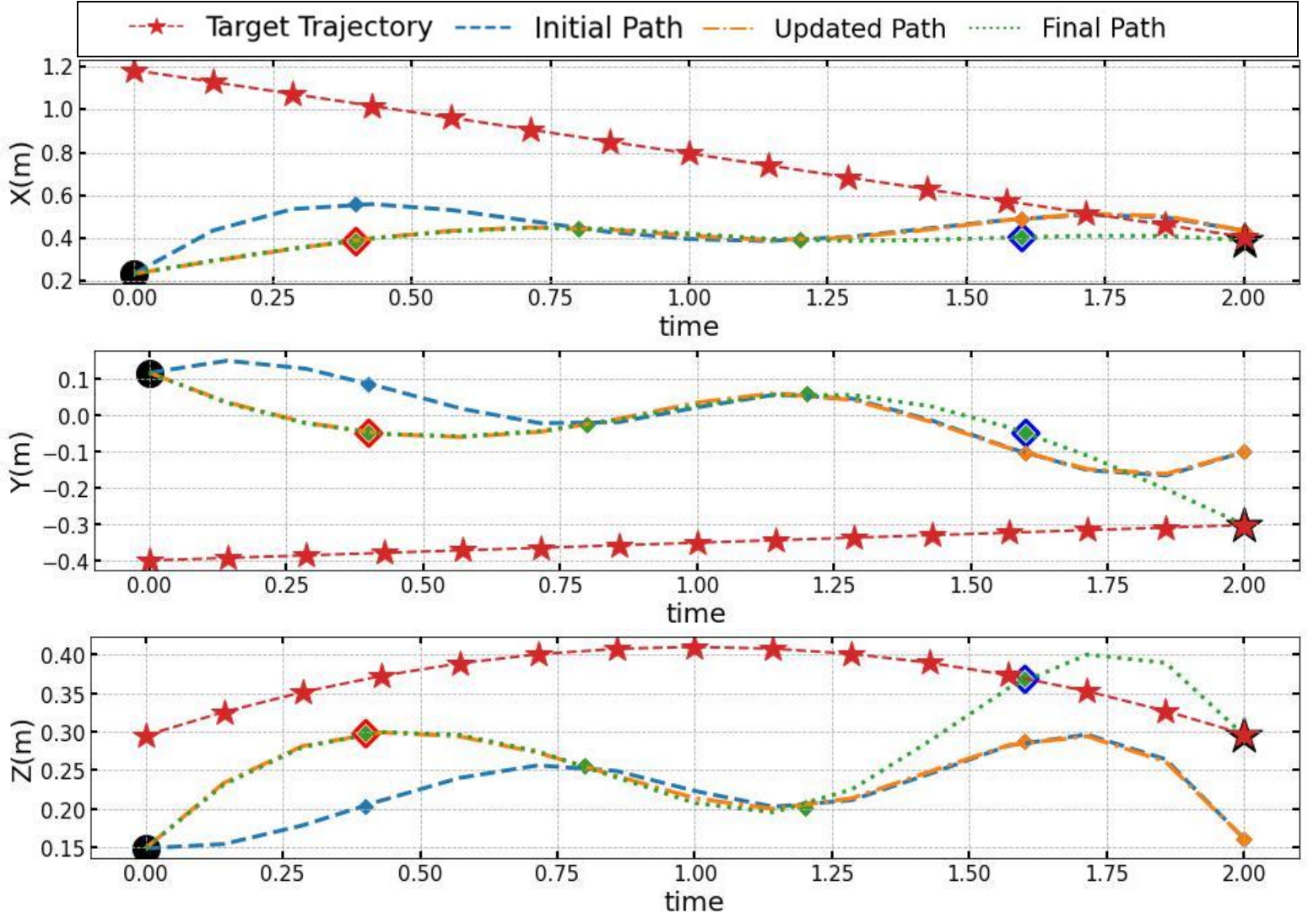}  
    \caption{The robot end-effector trajectory changes for moving obstacle location and updated intercepting location. Here, \textcolor{red}{$\Diamond$} represents the trajectory reroute location for obstacle point and \textcolor{blue}{$\Diamond$} denotes the time-step point where the intermediate path gets upgraded for updated intercepting point. $\star$ denotes the real target location for the thrown target object.}% 
    \label{fig:position}                                   
\end{figure}

We start to answer the first question by running computer simulations for intercepting flying objects under obstacle environments. Fig.~\ref{fig:software sim} gives an example illustration on how the robotic arm tries to capture the dynamic objects while replanning the routes so that it can avoid the potential collision. We observed that the robotic arm was lifted up to change the moving trajectory as it was close to the obstacles (Fig.~\ref{fig:software sim} c, d, and e), which ultimately intercept the objects without touching any obstacles (Figure~\ref{fig:software sim} f). Also, we observed that the replanned tracking trajectory is smooth without any rigid movement (Fig.~\ref{fig:position}). Further, our replanned tracking trajectory shows how the position of the robotic arm dodges the obstacles and reroutes through a new path to catch the moving target in three-dimensional space (see Fig.~\ref{fig:position}). The above simulation illustration example gives us the first confidence about the potential advantages of our proposed framework. Next, we tested our proposed framework on the hardware platform for intercepting a flying ball (Fig.~\ref{fig:platforms}). To emphasis the ability of real-time obstacle avoidance, we introduce the real-time obstacles on purpose during the interception tasks. It turned out that our robotic arm can react to the real-time obstacles by replanning a new collision-free trajectory, which finally intercepted the ball. As a result, both computer simulations and hardware evaluation examples gave us strong evidence about the superior performance of our proposed framework. 
% In the next few sections, we use both Monte Carlo simulations as well as the hardware platform to further evaluate our proposed framework.  

% \begin{figure}                                     
%     \centering                                             
%     \includegraphics[width=\linewidth]{./figures/compile.pdf}  
%     \caption{\textcolor{red}{Need to change a bit, emphasis on the more on the replanning part instead of the whole trajectory. we can discuss it on Monday }}% 
%     \label{fig:hardware sim}                                   
% \end{figure} 

\subsection{Comparison with the state-of-the-art robotic motion planning algorithms}

\begin{figure}[htbp]                                       
    \centering                                             
    \includegraphics[width=\linewidth]{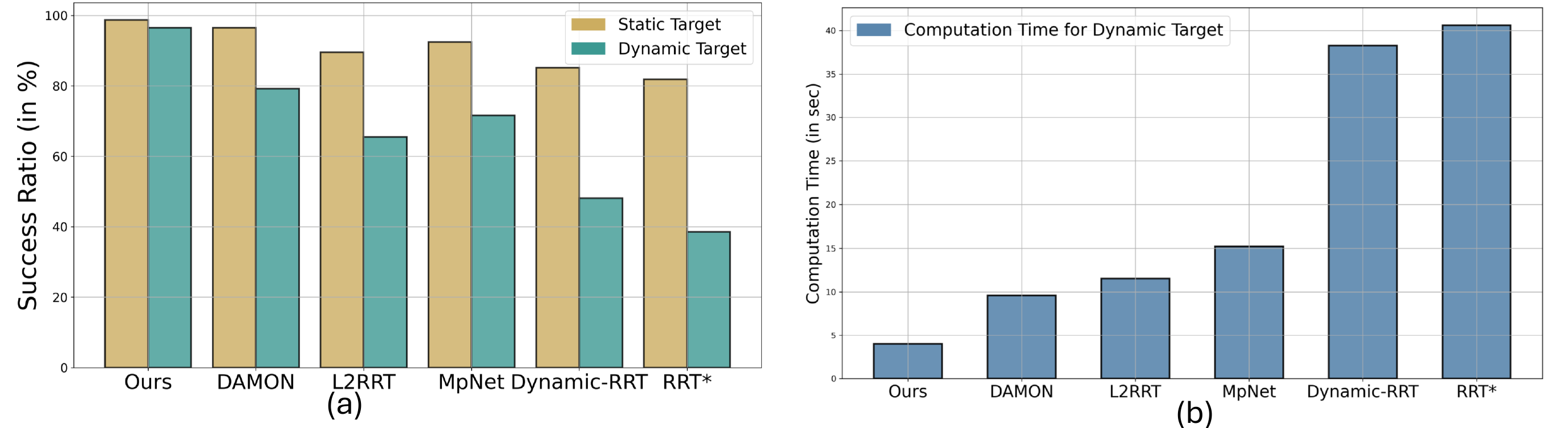}  
    \caption{(a) Successful ratio comparison between different state-of-the-art robotic motion planning algorithms for both tracking static objects and flying objects. The results for static objects are in green and the flying objects are in blue. (b) Computation time comparison for dynamic target interception.}
    \label{fig:Success_Ratio_comparison_SOTA}
\end{figure}

In this section, we compare our proposed framework with other robotic motion planning algorithms for the task of intercepting flying objects. Here, we focus on the successful ratio of catching flying objects and compare it with our previous proposed method DAMON as well as other methods such as MpNet, L2RRT and RRT variants. We start with simple cases where we ask the robotic arms to track a static target. Since the target is stationary over time, all the planning algorithms only need to avoid the obstacles considering the possible collision during the motion planning process. As a result, all the successful ratios are more than $80\%$, where our proposed method and our prior work DAMON achieved more than $95\%$. We argue this superior performance is due to the power of dimensionality reduction using the neural network architecture of variational autoencoder (see next section for more detailed analyses of D-VAE). On the contrary, several RRT variants calculate the trajectory in the original high-dimensional space which increases the planning difficulties and drops in the successful ratio (see Figure~\ref{fig:Success_Ratio_comparison_SOTA}(a)). Next, we move the tracking performance for flying objects. It was obvious that all the RRT-based methods can not achieve good successful ratios, which dropped below $70\%$. Our prior work DAMON did the best among the SOTA algorithms, which was almost $80\%$. However, our proposed method achieved more than $94\%$ successful ratio, which is significantly higher than SOTA algorithms. Compared to the DAMON, we argue the great improvement may be caused by our real-time object estimation mechanisms, i.e., use EKF to calculate nonlinear trajectory prediction, which reduces potential collision in advance. Moreover, D-VAE generates very smooth latent space in $\mathbb{R}^2$ which allows for fast and efficient motion planning computation through obstacle-free closest points in joint-space. In Fig.~\ref{fig:Success_Ratio_comparison_SOTA}(b), we showed that the computation time for ours proposed method for successful motion planning is the lowest among all other methods.
% A snapshot comparison of different robotic motion planning algorithms for tracking flying objects can be found in our online videos (\textcolor{red}{Please add in the online video}). 
The above experimental results confirmed that our proposed framework can lead to great improvement in tracking both static and dynamic objects while avoiding obstacles in real time.         

\subsection{Two-Dimensional embedding using D-VAE and full embedding graph traversal}
\begin{figure}                                      
    \centering                                             
    \includegraphics[width=\linewidth]{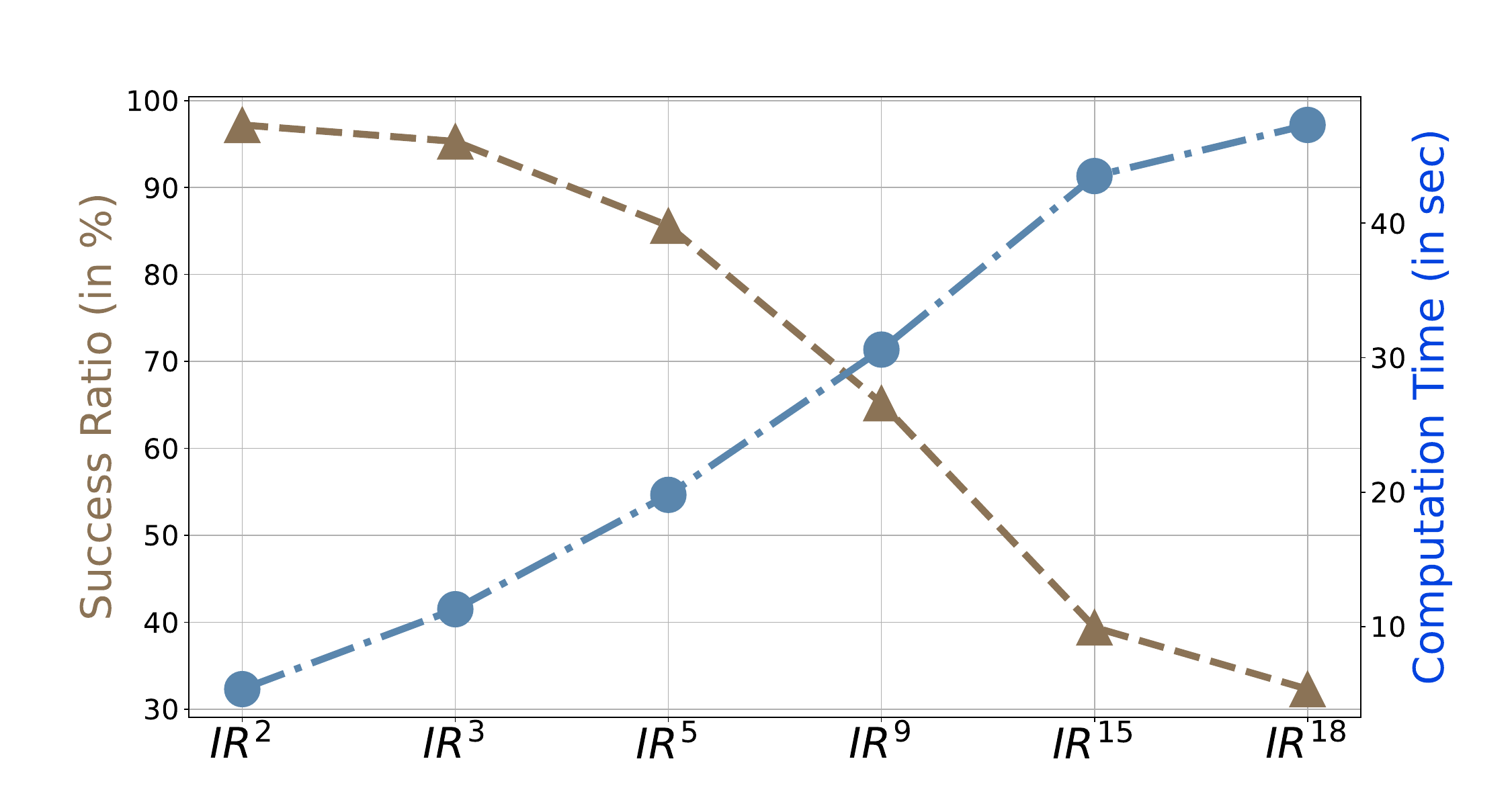}  
    \caption{Comparison of the successful ratio (light brown) and computation time (blue) using different embedding dimensions.} 
    \label{fig:Accuracy_&_computationTime_comparison_embedding_dimension}
\end{figure}

The prior section argues the improvement of the successful task completion is partly due to the power of D-VAE for dimensionality reduction and smooth latent space. Especially, we can easily build the graph in lower dimensions and use graph traversal algorithms for trajectory planning with lower computational cost. From high-dimensional data samples of $18$ numeric values, we use D-AVE for creating the representation of embedding manifold in $\mathbb{R}^2$ space. Therefore, we run the first set of ablation studies by comparing different embedding dimensions of our proposed algorithms. Fig.~\ref{fig:Accuracy_&_computationTime_comparison_embedding_dimension} shows that our methods achieved higher successful ratio with lower computational time for graph traversal in $\mathbb{R}^2$ space compared to the case in which it operated in $\mathbb{R}^{18}$ space. 
(Successful ratio: $\mathbb{R}^2$  97.23\% v.s. $\mathbb{R}^{18}$  32.38\% and Computational time: $\mathbb{R}^2$  5.36s v.s. $\mathbb{R}^{18}$  47.61s). Further, we consistently observed that the lower dimension ensures less computational time and higher successful ratio across a wide range of embedding dimensions. The above study confirmed that using D-AVE can improve the successful ratio and greatly decrease the computational cost. 

\begin{figure}                                      
    \centering                                             
    \includegraphics[width=\linewidth]{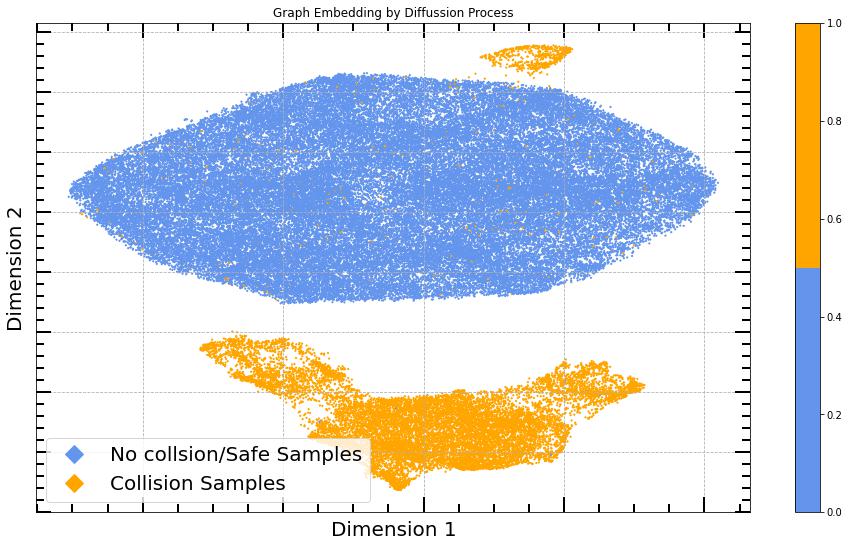} 
    \caption{Low dimensional embedding of ``collision-free'' samples (in blue color) or ``colliding'' samples (in orange color). }
    \label{fig:Low dimensional embedding of two samples}
\end{figure}

In detail, our D-VAE can effectively segregate between two binary labels of either ``collision-free'' samples or ``colliding'' samples, which greatly capture the geometric properties of the interaction of robotic systems and environment obstacles (see Fig.~\ref{fig:Low dimensional embedding of two samples}). To visualize the trajectory generation process when our original path anticipates a potential obstacle, we provide the following illustration example in Fig.~\ref{fig:dynamic_adaption}. We observed our methods automatically avoid the obstacle with almost the minimum change of the original path, which partially explains the lower computational time (Fig.~\ref{fig:dynamic_adaption}). The smooth latent space allows our method to adaptively re-route through the closest nodes in graph to reach new target location while avoiding any possible collisions.  

\subsection{Real-time object estimation using extended Kalman filter}
% \textcolor{red}{I think figure 11 and figure 12 can combine together, like what you did for successful ratio and computational time}
Having established the power of D-VAE, we lastly illustrate the importance of the extended Kalman filter (EKF). Using our proposed EKF for the real-time estimation of the flying objects (see section~\ref{Extended Kalman filter (EKF) for real-time estimation of the moving object} for details), we realize a precise position estimation of the objects. 
% Figure~\ref{fig:EKF Prediction} provides an illustration example of how our proposed EKF corrected continuously the previous estimation trajectory in real time. 
With new target locations, the target node in the graph is also updated for better interception. We observed that without EKF, the trajectory tracking deviation gradually accumulated, which finally led to a huge deviation and possibly caused the failure of intercepting the flying target objects. On the contrary, the EKF can modify the tracking trajectory by executing the real-time estimation of target objects, which successfully tracks the oracle moving trajectory with smallest tracking errors. 
% (\textcolor{red}{PUT number here}) The numbers are added later. 
To further confirm our observations, we run comprehensive experiments and also compare our EKF with other linear and polynomial estimation algorithms such as linear/polynomial regression, and nonlinear estimation algorithms such as the B-spline algorithm. First, we observed that using static estimation algorithms causes large estimation errors as they only use current position information to predict. On the contrary, the EKF has nonlinear and recursive mechanisms, which utilize both current and past position information for the prediction of future object locations. In detail, our EKF achieved cumulatively $0.67$ estimation errors, which outperformed other algorithms OLSReg (Ordinary Least Squares Linear Regression): $6.37$; PloyReg (Polynomial Regression): $4.25$; B-Spline: $2.81$. Therefore, by integrating EKF, our method achieved the higher success ratio $98.6\%$, which significantly outperformed other algorithms OLSReg: $15.6\%$; PloyReg: $52.3\%$; B-Spline: $71.8\%$. The above ablation study confirmed that the real-time nonlinear and recursive estimation of the target moving object is necessary for improving the final successful ratio.  
      
% \begin{figure}                                      
%     \centering                                             
%     \includegraphics[width=1\linewidth]{./figures/int.png}  
%     \caption{
%         The trajectory estimation for the position of the target object} 
%     \label{fig:EKF Prediction}                                   
% \end{figure}

% \begin{figure}                                      
%     \centering                                             
%     \includegraphics[width=1\linewidth]{./figures/Norm_comparison_on_estimation_method.pdf}  
%     \caption{Estimation error using different estimation methods}
%     \label{fig:Estimation error using different estimation methods}         
% \end{figure}

\begin{figure}                                      
    \centering                                             
    \includegraphics[width=1\linewidth]{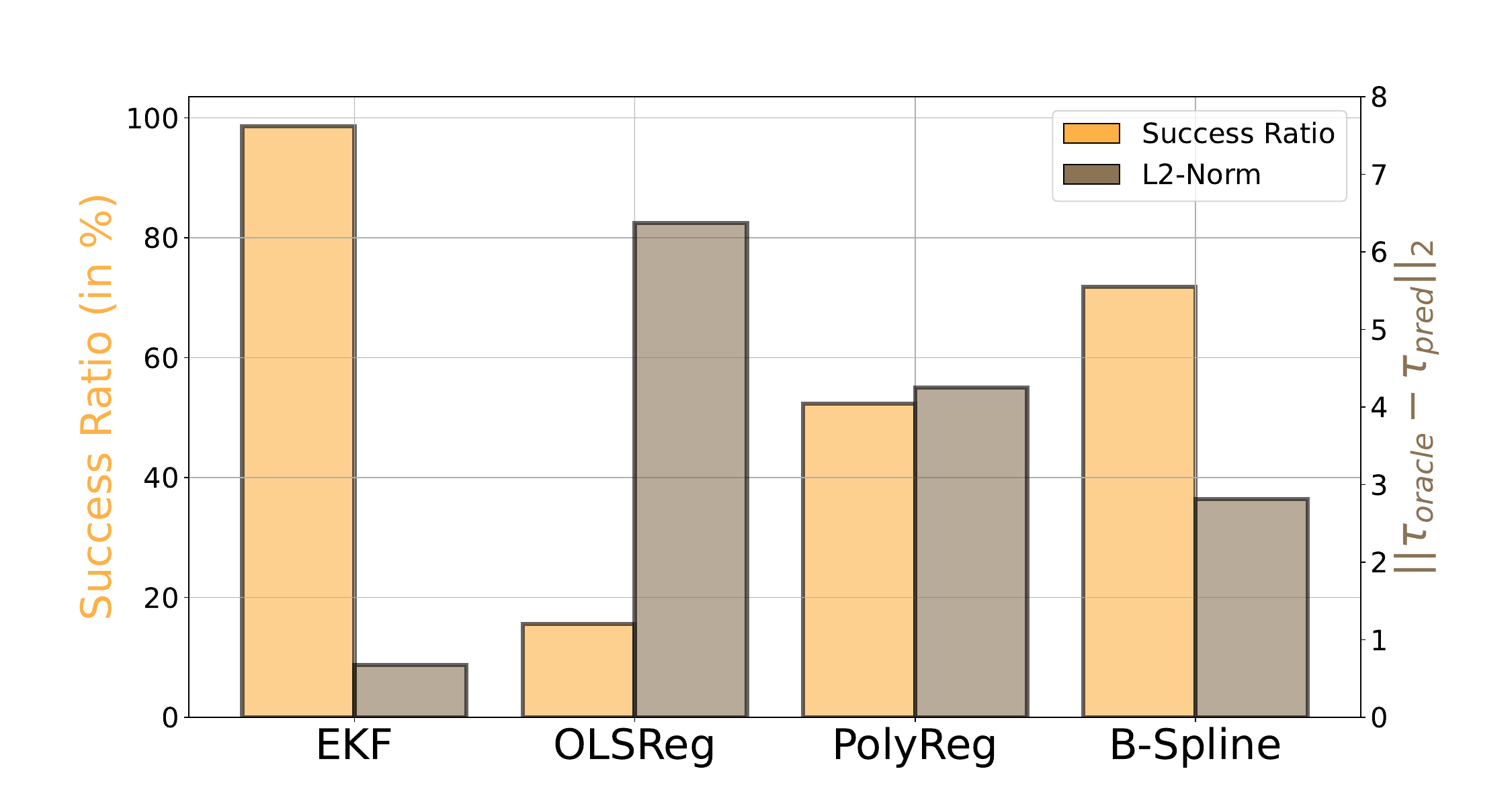}  
    \caption{Successful ratio using different estimation methods} 
    \label{fig:Success_Ratio using different estimation methods}            
\end{figure}

\section{Discussion and Conclusion}
In our work, we propose a unified control framework using diffusion variational autoencoder (D-VAE) for intercepting flying objects with dynamic obstacles. The key advantage of using D-VAE is the feasibility of dimensionality reduction from the original $18$ dimensional space to a two-dimensional manifold, which learns lower dimensional embedding and efficiently reduces the redundant robotic manipulator parameters. On the two-dimensional embedding, we construct a dense-connected graph and develop a path routing algorithm, which enables the fast synthesis of control commands in real time. Future work can incorporate more advanced graph traversing algorithms for improving the path routing efficiency with minimum energy cost requirement~\cite{Zhang2023FSGraphFA}. Our work borrows the EKF for the real-time position prediction of target objects. Using more advanced Kalman filters from the adaptive dynamical system approaches is another future direction~\cite{fang2022designing}. We demonstrate the effectiveness of our proposed algorithms on both computer simulations and autonomous 7-DoF robotic arms. Our results hold promise for the future generalization of robotic motion planning algorithms to handle dynamic moving obstacles while intercepting a moving target in real time. Giving the generalizablity of our proposed framework, we envision future validation on more complicated robotic systems such as dexterous in-hand manipulation~\cite{openai_dex_hand} and robotic dogs~\cite{zhao_legged_robot}.

\bibliographystyle{./bibliography/IEEEtran}
\bibliography{bib}

\end{document}